# A dynamic graph-cuts method with integrated multiple feature maps for segmenting kidneys in ultrasound images

**Qiang Zheng**
*Department of Radiology, School of Medicine, University of Pennsylvania, Philadelphia, PA, 19104, USA; School of Computer and Control Engineering, Yantai University, Yantai, 264005, China*
**Steven Warner and Gregory Tasian**
*The Children's Hospital of Philadelphia, Philadelphia, PA, 19104, USA*
**Yong Fan[a]**
*Department of Radiology, School of Medicine, University of Pennsylvania, Philadelphia, PA, 19104, USA*
[a] *Corresponding author: E-mail: yong.fan@ieee.org*

**Purpose:**

To improve kidney segmentation in clinical ultrasound (US) images, we develop a new graph cuts based method to segment kidney US images by integrating original image intensity information and texture feature maps extracted using Gabor filters.

**Methods:**

To handle large appearance variation within kidney images and improve computational efficiency, we build a graph of image pixels close to kidney boundary instead of building a graph of the whole image. To make the kidney segmentation robust to weak boundaries, we adopt localized regional information to measure similarity between image pixels for computing edge weights to build the graph of image pixels. The localized graph is dynamically updated and the GC based segmentation iteratively progresses until convergence. The proposed method has been evaluated and compared with state of the art image segmentation methods based on clinical kidney US images of 85 subjects. We randomly selected US images of 20 subjects as training data for tuning the parameters, and validated the methods based on US images of the remaining 65 subjects. The segmentation results have been quantitatively analyzed using 3 metrics, including Dice Index, Jaccard Index, and Mean Distance.

**Results:** Experiment results demonstrated that the proposed method obtained segmentation results for bilateral kidneys of 65 subjects with average Dice index of 0.9581, Jaccard index of 0.9204, and Mean Distance of 1.7166, better than other methods under comparison ($p<10^{-19}$, paired Wilcoxon rank sum tests).

**Conclusions:** The proposed method achieved promising performance for segmenting kidneys in US images, better than segmentation methods that built on any single channel of image information. This method will facilitate extraction of kidney characteristics that may predict important clinical outcomes such progression chronic kidney disease.

# 1. INTRODUCTION

Ultrasound (US) imaging has been widely used to examine for structural abnormalities of the kidney and to ascertain features such as renal parenchymal area that have been associated with development of end stage renal disease (Cost, Merguerian et al. 1996, Pulido, Furth et al. 2014, Gao, Perlman et al. 2017). However, automatic segmentation of kidneys in ultrasound (US) images remains a challenging task due to high speckle noise and low contrast between foreground and background, as well as weak boundaries and large appearance variations of kidneys in US images (Noble and Boukerroui 2006). Automatic segmentation of ultrasound images of kidneys will facilitate extraction and quantification of features such as renal parenchymal area and kidney echogenicity, which currently are measured manually or are subjective assessments, respectively.

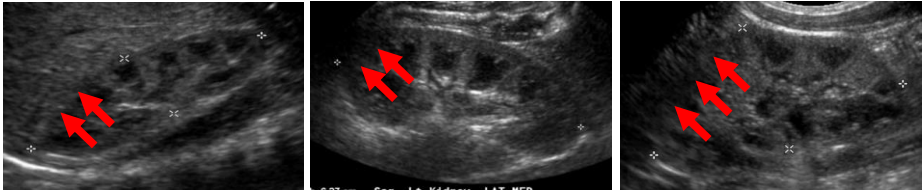

**Fig.1** Typical Kidney US images with speckle noise, low contrast between foreground and background, and weak boundaries. All the kidney images have large appearance variations within the kidneys.

A variety of methods have been proposed for segmenting kidneys in US images, including active contour model (ACM) based methods (Yang, Qin et al. 2012, Song, Wang et al. 2016), atlas-based methods (Marsousi, Plataniotis et al. 2015), Markov random field based methods (Martin-Fernandez and Alberola-Lopez 2005), watershed based methods (Tamilselvi and Thangaraj 2012), and machine learning based methods (Ardon, Cuingnet et al. 2015). Among them, the ACM based method is appealing for its robustness to imaging noise, weak boundaries, and large appearance variation within kidneys. However, the existing ACM based image segmentation methods typically adopt gradient descent flow based optimization techniques, which often get stuck at local minima (Bae and Tai 2008). Such a limitation can be overcome by graph cuts (GC) techniques (Bae and Tai 2008, Liu, Tao et al. 2011, Tao 2012). Particularly, GC techniques model the image segmentation task as an image labeling problem on a graph (Boykov, Veksler et al. 2001, Xu, Ahuja et al. 2003, Boykov and Kolmogorov 2004, Le, Jung et al. 2010). The GC techniques can be integrated with the ACM based methods by transforming the minimization problem of the ACM methods into a min-cut problem of a graph (Xu, Ahuja et al. 2003, Bae and Tai 2008, Liu, Tao et al. 2011, Tao 2012, Zhang, Liang et al. 2013, Tian, Liu et al. 2016). However, speckle noise and low contrast of US images might degrade the segmentation performance if the US images are directly used as input to the segmentation algorithms.

Texture analysis has been adopted to characterize US images in image segmentation studies and led to better performance than image segmentation based on image intensity information alone (Xie, Jiang et al. 2005, Faucheux, Olivier et al. 2012, Jokar and Pourghassem 2013, Cerrolaza, Safdar et al. 2016). Texture feature extraction techniques can be categorized as structural (Mehrnaz and Jagath 2009, Hacihaliloglu, Abugharbieh et al. 2011, Jia, Mellon et al. 2016), statistical (Wu, Gan et al. 2015), model-based (Cohen and Cooper 1987, Xie, Jiang et al. 2005), and transform-based methods (Haralick, Shanmugam et al. 1973, Daugman 1985, Malik, Belongie et al. 2001, Cunha, Zhou et al. 2006, Jokar and Pourghassem 2013, Cerrolaza, Safdar et al. 2016). In particular, structural methods utilize geometric primitives to represent textures, statistical methods represent texture features by statistics of intensity distributions or relationships among them, and model-based methods generally assume that texture features obey a certain statistical distribution. Different from aforementioned methods, transform-based methods are typically built upon multi-scale frequency and multi-direction analysis and have been demonstrated capable of capturing texture information more effectively (Daugman 1985). Particularly, Gabor transformation (Daugman 1985, Sandberg, Chan et al. 2002, Cerrolaza, Safdar et al. 2016) is a representative transform based method for characterizing image texture information. A Gabor transformation consists of a set of filters, which are multiplications of a Gaussian function and a sinusoidal function in different directions and frequencies, capable of capturing texture information similar to perception in the human visual system (Marcelja 1980, Daugman 1985, Jones and Palmer 1987). Although many US image segmentation studies have adopted texture information in the image segmentation (Xie, Jiang et al. 2005, Faucheux, Olivier et al. 2012, Jokar and Pourghassem 2013, Cerrolaza, Safdar et al. 2016), most of them only consider a single texture feature map for image segmentation, which might be insufficient to deal with imaging noise of clinical data.

Building upon the success of existing image segmentation methods (Boykov, Veksler et al. 2001, Xu, Ahuja et al. 2003, Boykov and Kolmogorov 2004, Bae and Tai 2008, Tao 2012), we propose a new graph cuts based method to segment kidney US images by integrating original image intensity information and texture feature maps extracted using Gabor filters. To handle large appearance variation within kidney images and improve computational efficiency, we build a graph of image pixels close to kidney boundary instead of building a graph of the whole image. To make the kidney segmentation robust to weak boundaries, we adopt localized regional information to measure similarity between image pixels for computing edge weights to build the graph of image pixels. The localized graph is dynamically updated and the GC based segmentation iteratively progresses until convergence. Our method has been evaluated based on clinical kidney US images of 85 subjects.

## 2. MATERIALS AND METHODS

Our image segmentation method is built upon graph cuts and texture analysis techniques (Daugman 1985, Boykov, Veksler et al. 2001, Boykov and Kolmogorov 2004). The image segmentation problem is solved using GC techniques by building a dynamically evolving graph of image pixels close to the boundary of kidneys. Both original image intensity information and multi-scale and multi-directional Gabor features are integrated into a localized regional similarity measure for measuring similarity between image pixels.

### 2.1 Texture feature map extraction

A new method is proposed to extract texture feature maps based on multi-scale and multi-direction analysis using Gabor transformation.

#### 2.1.1 Gabor transform

Gabor transform is a windowed Fourier transform (Daugman 1985), consisting of Gabor filters in different scales and directions. A 2D Gabor filter is the multiplication of a Gaussian kernel function and a sine wave function:

$$\begin{cases} g = \frac{1}{2\pi\sigma_x\sigma_y} exp\left(-\left(\frac{x'^2}{\sigma_x^2} + \frac{y'^2}{\sigma_y^2}\right)\right) exp\left(i\left(2\pi\frac{x'}{\lambda}\right)\right), \\ x' = xcos\theta + ysin\theta, y' = -xsin\theta + ycos\theta \end{cases} \quad (1)$$

where $\sigma_x$ and $\sigma_y$ are standard deviations of the Gaussian functions along axis $x$ and axis $y$ respectively, $\theta$ is the direction of the Gabor filter, wavelength $\lambda$ is the frequency factor, and the frequency $\omega = \frac{2\pi}{\lambda}$. A Gabor filter with a small $\lambda$ captures image information of high frequency and small scale, while image information of low frequency and big scale can be captured by a filter with a big $\lambda$. Gabor filters in different scales and directions can be integrated to capture multiscale image information.

### 2.1.2 Extraction of texture feature maps

Given an US image $I$, using Gabor filters we compute texture features $F_{i,j}, i = 1, \ldots, m, j = 1, \ldots, n$ at $m$ scales in $n$ directions with values $f_{i,j}(p)$ at pixel $p$ of $I$. To enhance kidney boundaries and suppress noise in US images, we propose a novel scheme to fuse filtered images with greater responses, which amounts to a filtering process along edges in US images, as schematically illustrated in Fig. 2. Particularly, given a pixel $p$ with filtered values $f_{i,j}(p)$ corresponding to $i = 1, \ldots, m$ scales and $j = 1, \ldots, n$ directions, a dominant direction at each scale is first determined as

$$g_i = \underset{j=1,\ldots,n}{\mathrm{argmax}} |f_{i,j}(p)|, i = 1, \ldots, m, \tag{2}$$

where the dominant direction at scale $i$ has the maximum absolute value, $|f_{i,g_i}(p)|$, among all directions. For every direction, we then compute the number of scales at which the direction under consideration is the dominant direction.

$$D_j = \sum_{i=1}^{m} \delta(|f_{i,j}(p)| - |f_{i,g_i}(p)|), j = 1, \ldots, n \tag{3}$$

where function $\delta(\cdot)$ is 1 at zero and 0 otherwise. Finally, we identify dominant directions across all scales as

$$\Omega = \left\{ j | D_j \geq 0.5 \cdot \underset{j=1,\ldots,n}{\max} D_j \right\}, \tag{4}$$

where dominant directions are identified as those with filtering responses greater than half of the maximal response values to balance robustness and redundancy of the dominant directions.

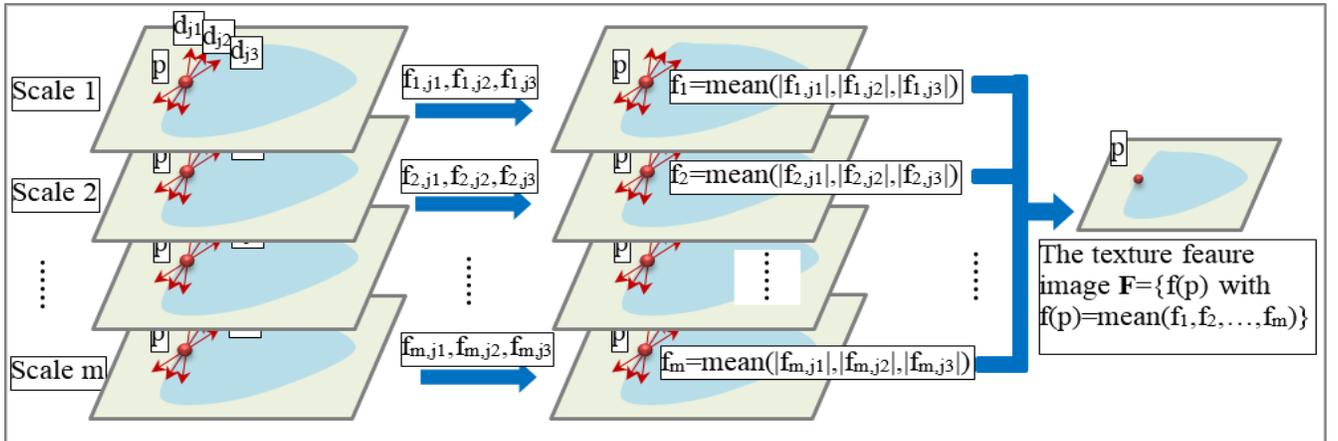

**Fig.2** Feature map extraction. Filtered images with greater responses at different scales and in different directions are fused to extract texture feature maps based on the Gabor transform.

After determining the multiple dominant directions (for instance, $d_{j1}, d_{j2}, d_{j3}$ as shown in Fig.2), we compute a texture feature map $\boldsymbol{F}$ with element $f(p)$ by fusing the filtered images in all dominant directions across all scales

$$f(p) = \frac{1}{m}\sum_{i=1}^{m}\sum_{j\in\Omega}|f_{i,j}(p)|. \tag{5}$$

Example Gabor feature maps are shown in Fig.3, along with their corresponding US images of 3 subjects. Both the US images and their corresponding Gabor features are to be used as feature maps in our image segmentation algorithm.

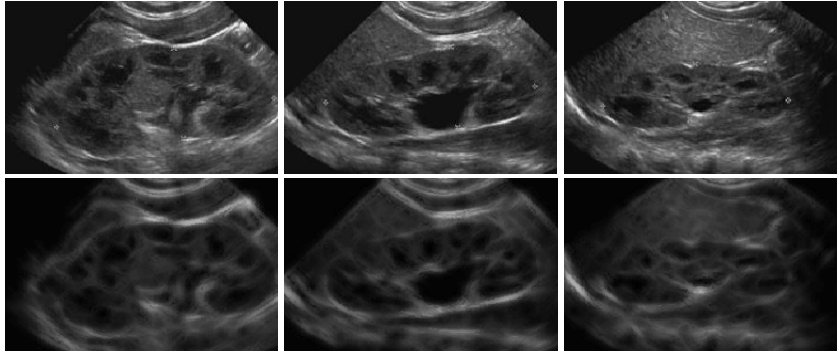

**Fig.3** Example feature maps of kidney US images. Top row: Original images; bottom row: Gabor feature maps.

**2.2 Dynamic GC based image segmentation with integrated multiple feature maps**

The GC based image segmentation methods model the image segmentation as a graph partition problem (Boykov and Jolly 2001). Given a set of pixels $\boldsymbol{P}$ to be segmented into a set of regions, the GC based segmentation methods model the pixels $\boldsymbol{P}$ as graph nodes $\boldsymbol{V}$ that are connected by edges $\boldsymbol{E}$ weighted by image similarity measures $\boldsymbol{W}$, i.e., the image is modeled as a graph $G = (\boldsymbol{V}, \boldsymbol{E}, \boldsymbol{W})$. A cut on the graph $G$ can be seen as a contour on the graph, which separates graph nodes $\boldsymbol{V}$ into two disjoint subsets (segments) $\boldsymbol{S}$ and $\boldsymbol{T}$, and the cost of the cut is measured by $|CUT_C|_G = \sum_{p\in S, q\in T} w(p,q)$, where $w(p,q)$ is a weight of the edge connecting pixels $p$ and $q$. The cut with the minimal cost (min-cut) can be obtained using min-cut/max-flow algorithms (Boykov and Kolmogorov 2004).

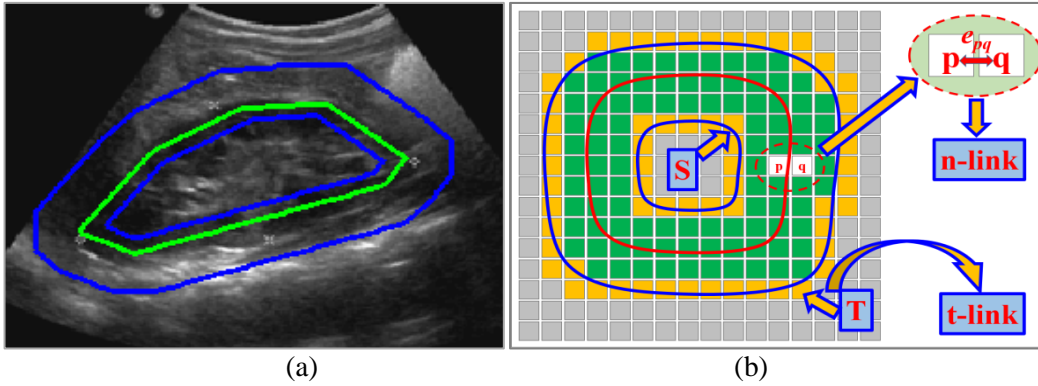

(a)                        (b)

**Fig.4** A graph of pixels within a narrow band of the kidney boundary. (a) A narrow band surrounding the kidney boundary: the green curve is an initialization of kidney boundary and a narrow band is located in-between the blue curves which are obtained by inflating and shrinking the green contour; (b) A graph of pixels within the narrow band (green pixels) in-between the blue cures (denoted by yellow pixels): n-links connect neighboring pixels $(p, q)$, and t-links connect yellow pixels inside or outside the green pixels with $S$ or $T$ respectively so that the pixels are segmented into disjoint subsets of $S$ or $T$.

The performance of GC based image segmentation is hinged on the graph of pixels to be segmented, particularly the weights of edges that connect the pixels. Since the kidney has large appearance variations and weak boundaries in US images, it is difficult to build a graph of all pixels of an US image to obtain satisfied segmentation performance. Motivated by the success of localized image segmentation method (Xu, Ahuja et al. 2003, Tao 2012), we build a graph of pixels surrounding the kidney boundaries and gauge similarity between pixels using a combination of pixel and regional measures based on the multiple feature maps, as illustrated by Fig. 4. Particularly, a narrow band surrounding the kidney boundaries is identified by inflating and shrinking an initialization of kidney boundary. The weight $w(p, q)$ for an edge connecting two pixels $p$ and $q$ is defined as an image similarity measure between the pixels:

$$w(p,q) = w_p(p,q) + w_r(p,q),$$
$$w_p(p,q) = exp\left(-\frac{\frac{1}{N}\sum_{i=1}^{N}(I_i(p)-I_i(q))^2}{\sigma}\right), \ w_r(p,q) = exp\left(-\frac{\left(1/\left(\frac{1}{N}\sum_{i=1}^{N}K_i(p,q)\right)\right)}{\sigma}\right), \tag{6}$$

where $p$ and $q$ are adjacent pixels, $w_p(p,q)$ and $w_r(p,q)$ are image similarity measures based on pixel information and regional information respectively, $I_i(p)$ and $I_i(q)$ are image intensity value of $p$ and $q$ for the $i$th feature map, $K_i(p,q)$ is an image difference measure based on regional information, $\sigma$ is a parameter, and $i = 1$, 2 corresponding to the original US image intensity and the Gabor feature map respectively. The numerator terms in both $w_p(p,q)$ and $w_r(p,q)$ are normalized to $[0,1]$. Particularly, $K_i(p,q)$ is defined as

$$K_i(p,q) = \min_{(l_p,l_q)\in\{(S,T),(T,S),(S,S)\ or\ (T,T)\}} \left(I_i(p) - f_i^{l_p}(p)\right)^2 + \left(I_i(q) - f_i^{l_q}(q)\right)^2, \tag{7}$$

where $I_i(p)$ and $I_i(q)$ are image intensity value of $p$ and $q$ for the $i$th feature map, $f_i^S(p), f_i^T(p), f_i^S(q), f_i^T(q)$, are intensity mean of pixels of a small neighborhood of $p$ and $q$ corresponding to segments **S** and **T** respectively as illustrated by Fig. 5, and $l_p \in \{S,T\}$ and $l_q \in \{S,T\}$ are possible segmentation labels of $p$ and $q$ to be updated according to the minimal value computed by Eqn. (7). The optimal segmentation labels of $p$ and $q$ should minimize the differences between their intensity values and the mean of their neighborhood corresponding to their segmentation labels. However, if the optimal segmentation labels of $p$ and $q$ are the same, we set $K_i(p,q) = \max_{(u,v)\in(V,V)} K_i(u,v)$ so that they will not be separated into different segments.

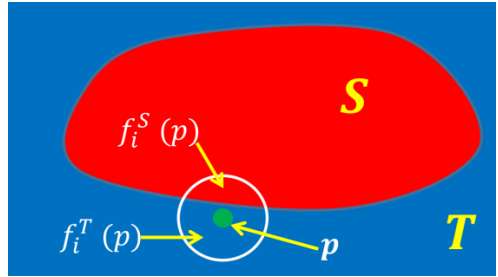

**Fig.5** Illustration of the computation of $f_i^S(p)$ and $f_i^T(p)$ given a segmentation result with segments **S** and **T**.

In the present study, a circular neighborhood system is adopted with radius $r = 10$ for computing intensity mean of pixels of a small neighborhood. As illustrated by Fig. 5, $f_i^S(p)$ and $f_i^T(p)$ of a pixel $p$ are intensity means of pixels within segments **S** and **T** respectively.

The edge weights and the segmentation result in the graph are updated iteratively until convergence. The segmentation algorithm is summarized as:

(1) Compute Gabor feature maps given US images and the numbers of scales and directions.

(2) Initialize a contour $C$ around but inside kidney boundary, and assign pixels with labels: $BW0(p) = \begin{cases} 1, \text{if } p \text{ is inside the contour,} \\ \quad 0, \text{otherwise.} \end{cases}$

(3) Build a narrowband of the contour $C$ by inflating and shrinking the initialization contour as illustrated by Fig. 4(a) with radii 15 and 3, respectively.

(4) Build the graph of pixels within the constructed narrowband as illustrated by Fig.4 (b):

n-links: pixels within the narrowband are connected with n-links weighted with $w(p,q) = w_p(p,q) + w_r(p,q)$ according to Eqn. (6);

t-links: pixels on the boundaries of the narrowband are connected with segments $S$ and $T$ weighted with $\infty$.

(5) Apply max-flow/min-cut algorithm (Boykov and Kolmogorov 2004) to the graph, and assign pixels with labels: $BW1(p) = \begin{cases} 1, \text{if } p \text{ is inside the contour,} \\ 0, \text{otherwise.} \end{cases}$

(6) Update the regional measure similarity $w_r(p,q)$ according the result contour.

(7) Repeat steps (4)-(6) until converge.

## 2.3 Segmentation performance evaluation and parameter optimization

The proposed method has been validated based on clinical kidney US images obtained from 85 subjects with their original image intensity information and Gabor feature maps as image features. The kidney US images were collected at the Children's Hospital of Philadelphia. Each subject's bilateral kidney images were manually segmented and treated as ground-truth for evaluating the automatic image segmentation performance.

For evaluating the image segmentation performance, we adopted Dice Index, Jaccard Index, and Mean Distance to quantitatively measure similarity/difference between the automatic segmentation results and manual labels. These quantitative metrics are defined as

$$\text{Dice Index} = 2\frac{V(E \cap F)}{V(E)+V(F)},$$

$$\text{Jaccard Index} = \frac{V(E \cap F)}{V(E \cup F)},$$

$$\text{Mean Distance} = \text{mean}_{e \in BE}\left(\min_{f \in BF} d(e,f)\right),$$

where $E$ and $F$ are two different segmentation results, $d(\cdot,\cdot)$ is Euclidian distance between two points, and $BE$ and $BF$ are boundary voxels of $E$ and $F$ respectively.

We randomly selected US images of 20 subjects to optimize parameters of the proposed method and evaluate how the parameters affect the segmentation performance. Based on the optimization parameters, we validated the algorithm based on US images of the remaining 65 subjects. We also compared our method with state of the art

US image segmentation methods, including localizing region based active contours (Lankton and Tannenbaum 2008), geodesic active contours (Caselles, Kimmel et al. 1997), and the adaptive multi-feature segmentation model (AMFSM) (Zhang, Han et al. 2016). The same initialization contours were used by all the methods under comparison.

## 3. RESULTS

### 3.1 Optimization of the parameters

The proposed method has several parameters, including the numbers of scales and directions for computing the Gabor feature maps, and $\sigma$ for computing the image similarity measures. A grid searching was adopted to select an optimal setting for the parameters based on US images of 20 subjects. In particular, the number of scales was selected from {2,3,4,5}, the number of directions was selected from {4,8,16}, and $\sigma$ was selected from {1, 0.1, 0.01}. Fig.6 shows plots of mean values of Dice Index and Mean Distance for segmentation results obtained at different settings of the parameters. The experiment results indicated that the best kidney segmentation performance (Dice Index >0.947, Mean Distance <2.06) could be obtained with the number of scales=3, the number of directions=8, and $\sigma = 0.1$. This setting was adopted in all following experiments unless otherwise specified.

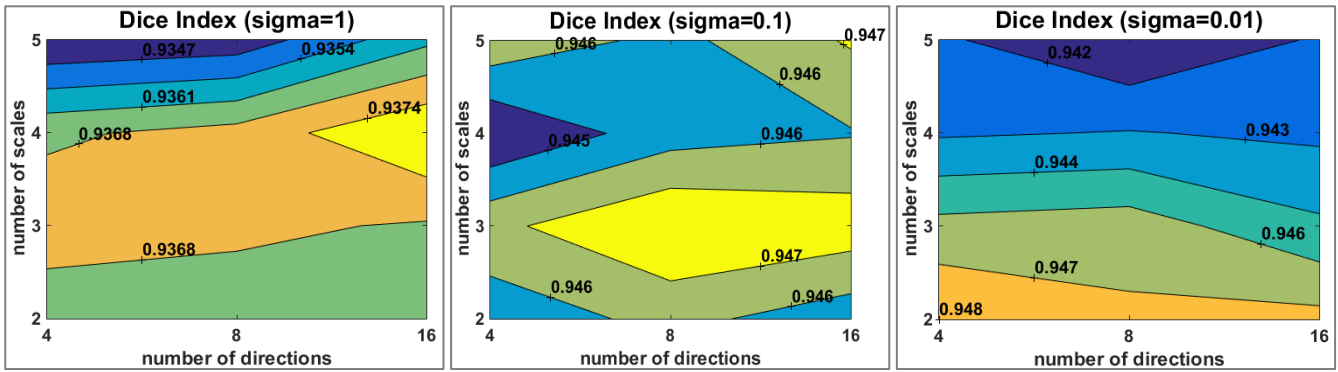

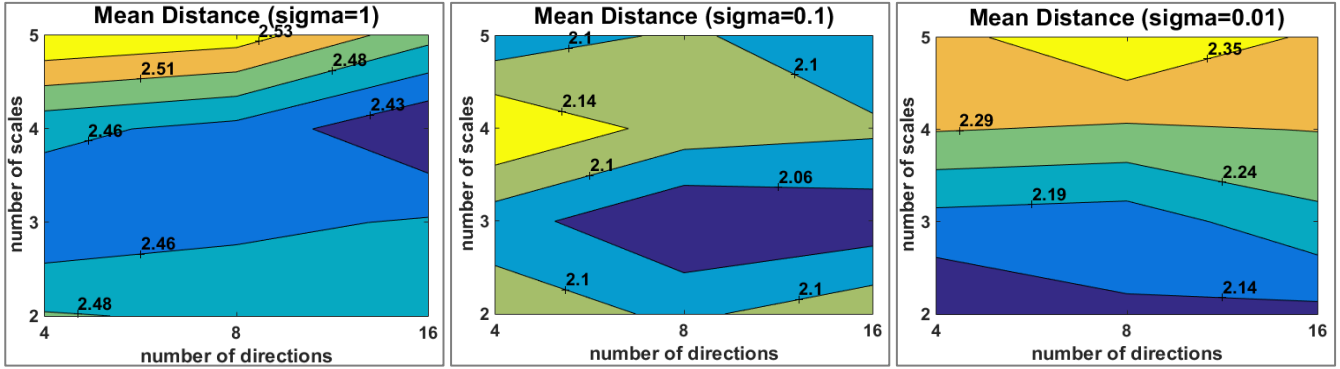

**Fig.6** Segmentation accuracy measures obtained with different numbers of scales and directions and different values of $\sigma$.

### 3.2 Experiments on single/multiple feature maps

Our method has been evaluated with different setting of image features, including segmenting images based on the original images, the Gabor feature map, and their combination based on US images of 20 subjects. The kidney segmentation based on either the original image intensity information or the Gabor feature map alone is a degraded version of our proposed method. The same grid searching scheme was adopted to optimize parameters of the kidney segmentation based on the single feature map. Particularly, the best kidney segmentation performance based on the image intensity information was obtained with $\sigma = 0.1$(summarized in Table 1) , and the best kidney segmentation performance based on the Gabor feature maps (Dice Index >0.943, Mean Distance <2.21) was obtained with the number of scales=2, 3, the number of directions=4, 8, 16, and $\sigma = 0.1$ as shown in Fig. 7. To evaluate the kidney segmentation method based on the single feature map, the same parameters as those used in the proposed multi-feature maps based segmentation method, i.e., the number of scales=3, the number of directions=8, and $\sigma = 0.1$.

**Table 1** Segmentation accuracy measures obtained with different values of $\sigma$.

|  | Dice | Mean Distance |
|---|---|---|
| $\sigma = 1$ | 0.9408 | 2.3792 |
| $\sigma = 0.1$ | 0.9466 | 2.0713 |
| $\sigma = 0.01$ | 0.9356 | 2.3179 |

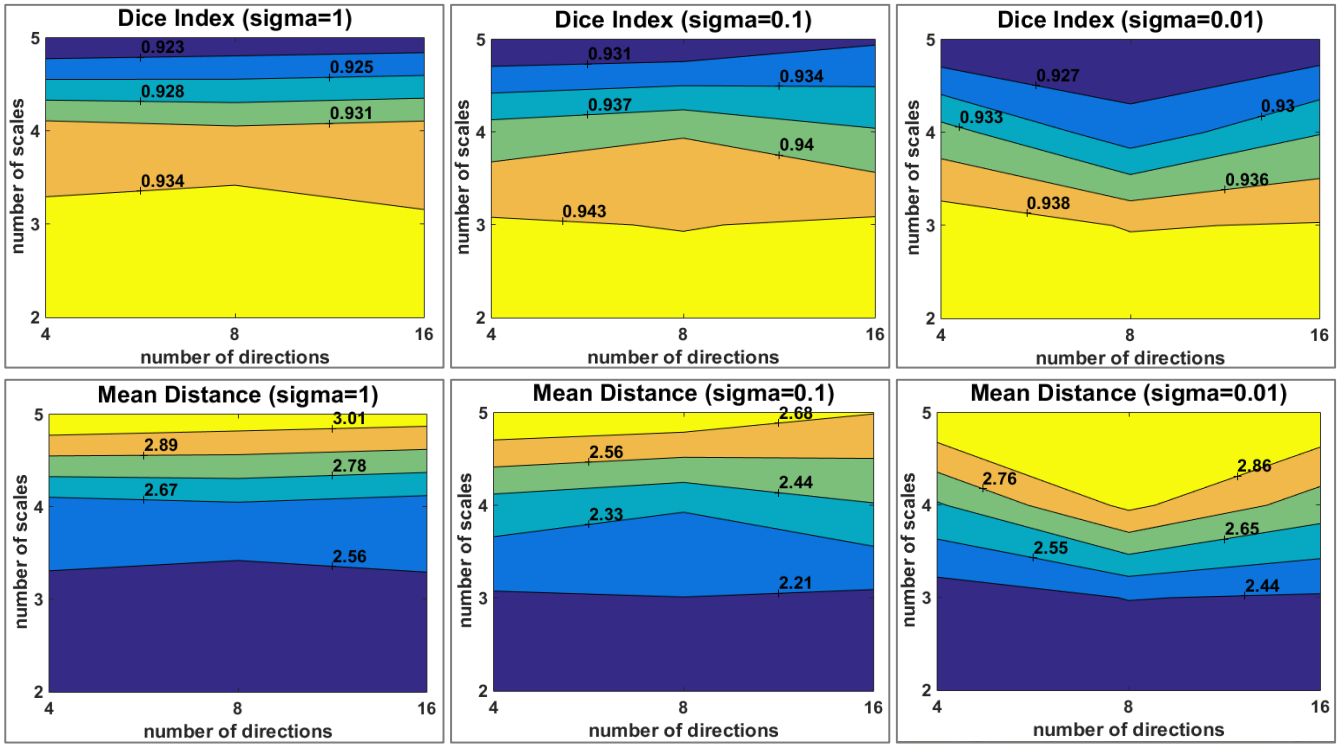

**Fig.7** Segmentation accuracy measures obtained with different numbers of scales and directions and different values of $\sigma$.

Box plots of segmentation accuracy measures of single/multiple feature maps are shown in Fig. 8, and segmentation results of an example US image are shown in Fig. 9. These results demonstrated that the proposed multiple feature maps based method could achieve better performance than the alternatives. Therefore, the setting of multiple feature maps was adopted in all following experiments.

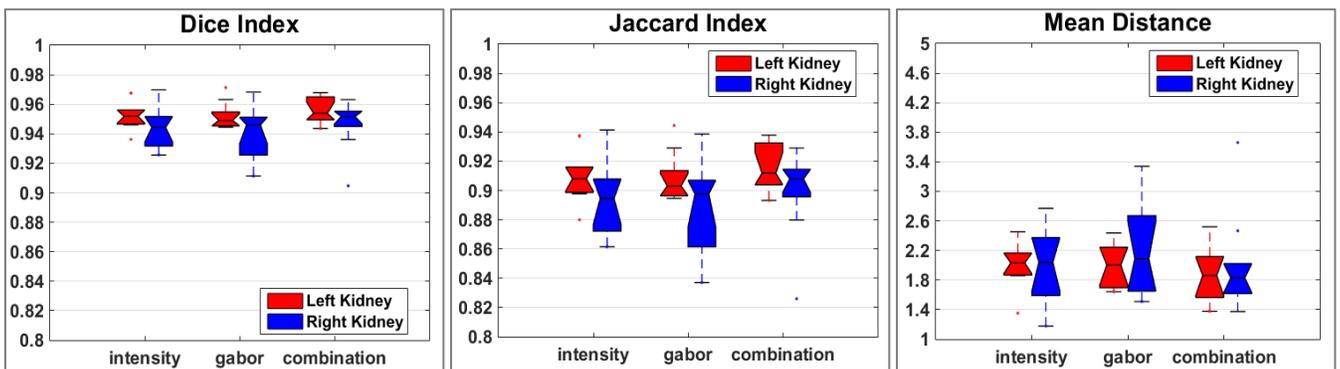

**Fig.8** Box plots of kidney (left and right) image segmentation accuracy measures of 20 subjects based on the original image intensity, the Gabor feature map, and their combination.

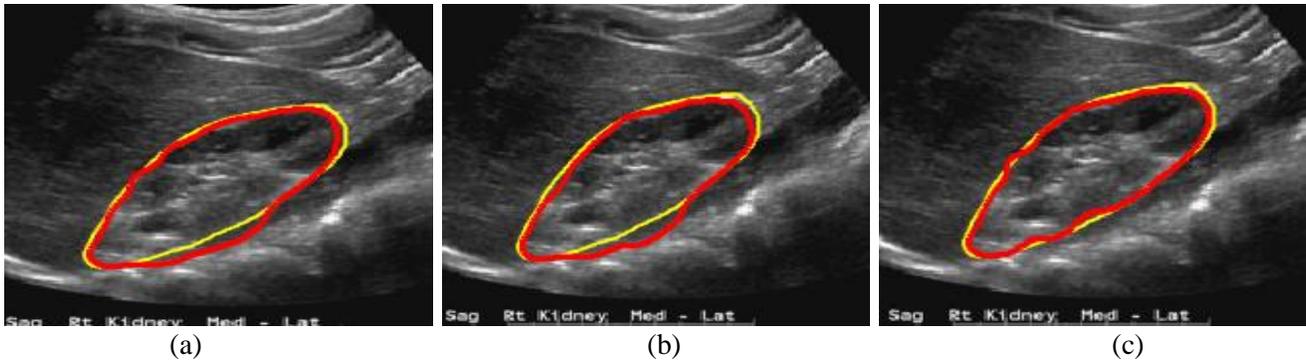

(a)                        (b)                        (c)

**Fig.9** Kidney segmentation on US images. Red curves are segmentation results obtained by the automatic segmentation algorithm with different feature maps, and yellow curves are manual segmentation results. (a) Segmentation result obtained based on the original image intensity information; (b) Segmentation result obtained based the Gabor feature maps; and (c) segmentation results obtained based on their combination.

### 3.3 Experiments on different settings of image similarity measures

The image similarity measure was defined based on the combination of both pixel information and local regional information. We evaluated how the image similarity measure affected the image segmentation performance. Fig.10 shows segmentation accuracy measures of US images of 20 subjects obtained with the image similarity measures defined based on the pixel information, the regional information, and their combination. Example segmentation results obtained by different settings are shown in Fig. 11. These experiments demonstrated that the image similarity measure defined based on the combination of both the pixel information and the regional information yielded the best kidney segmentation performance.

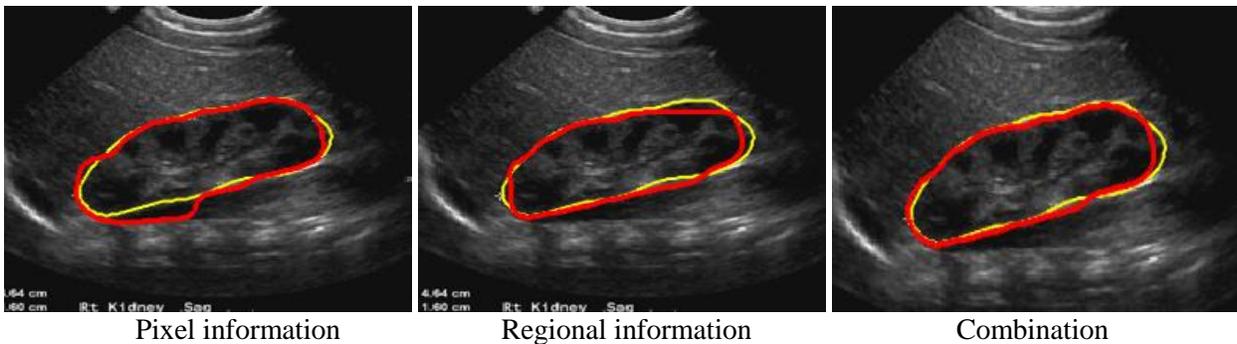

       Pixel information          Regional information          Combination

**Fig.10** Comparison experiments utilizing only pixel information, only regional information, and their combination. Automatic segmentation results are shown in red, and manual label results are shown in yellow.

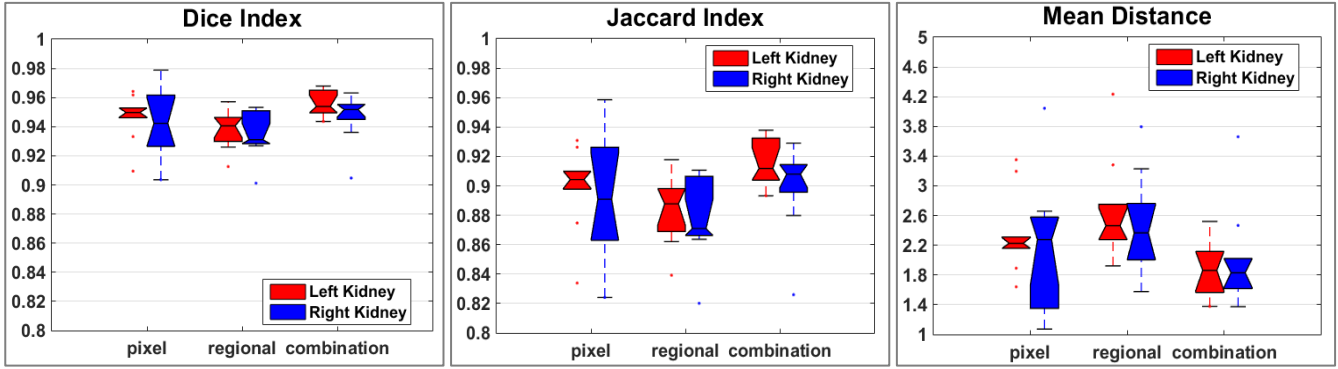

**Fig.11** Box plots of segmentation performance measures of 20 subjects, utilizing only pixel information, only regional information, and their combination.

### 3.4 Experiments on different initializations

The proposed method needs to have an initialization contour which could be obtained by manually picking 6-10 points to outline the shape of the kidneys. Fig.12 shows example initialization contours for 3 US images. To evaluate how the initialization affects the segmentation performance, we obtained segmentation results of US images of the 20 subjects with 3 different initialization contours. Quantitative segmentation results as shown in Fig. 13 indicated that the segmentation performance was relatively stable with respect to the initialization contours. The degree of consistency of the segmentation performance measures obtained with different initializations was further measured by intraclass correlation coefficients (ICCs, the segmentation performance measures was modeled as averages of k independent measurements on randomly selected objects) (McGraw and Wong 1996), as summarized in Table 2. The ICC results indicated that the proposed method was relatively robust to different initializations.

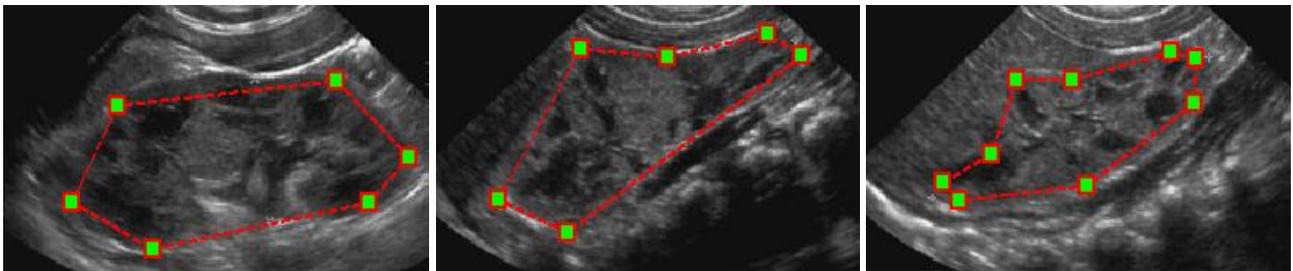

**Fig.12** Initialization examples. Red curves are initialization contours determined by the points in green.

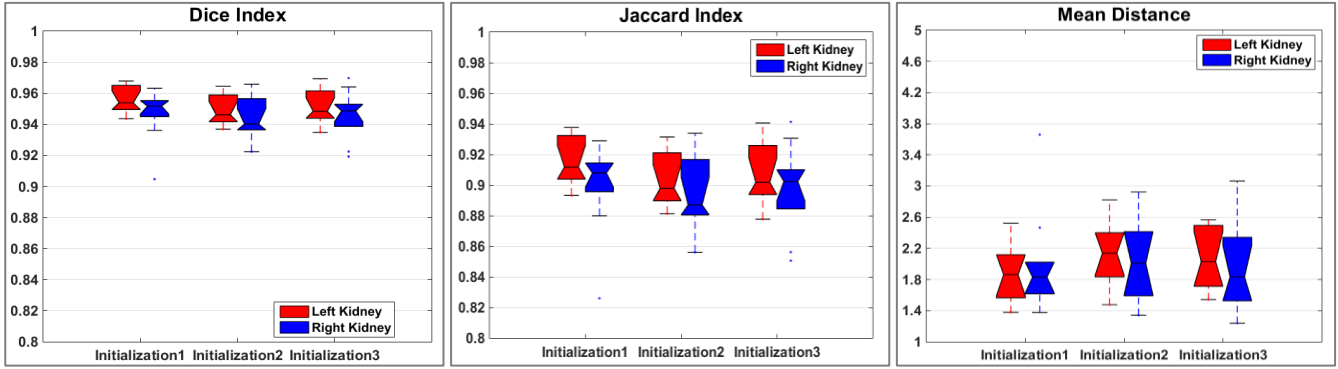
**Fig.13** Box plots of kidney image segmentation results for 20 subjects with 3 different initializations.

**Table 2**. Intraclass correlation coefficients for 3 different initializations

| Dice Index | Jaccard Index | Mean Distance |
|---|---|---|
| 0.89 | 0.89 | 0.90 |

**3.5 Comparison with state of the art methods**

We compared the proposed method based on US images of 65 subjects with state of the art image segmentation methods, including localizing region based active contours (LRGC) (Lankton and Tannenbaum 2008) and geodesic active contours (GAC) (Caselles, Kimmel et al. 1997). In particular, the implementation of LRGC and GAC was based on a software package "creaseg" (Dietenbeck, Alessandrini et al. 2010) that integrated 6 different image segmentation methods (Caselles, Kimmel et al. 1997, Lankton and Tannenbaum 2008, Li, Kao et al. 2008, Shi and Karl 2008, Bernard, Friboulet et al. 2009, Chan and Vese 2010), and LRGC and GAC methods were among the best. Moreover, a recent adaptive multi-feature segmentation model (AMFSM) (Zhang, Han et al. 2016) was also adopted in comparison. The same initialization contours were used for all the methods.

Fig.14 shows box plots of segmentation performance measures for results obtained by different methods, and the quantitative segmentation performance measures are summarized in Table 3, indicating that the proposed method performed significantly better than the alternatives ($p<10e^{-19}$, paired Wilcoxon rank sum tests). The experiment results demonstrated that the proposed method achieved better performance than others.

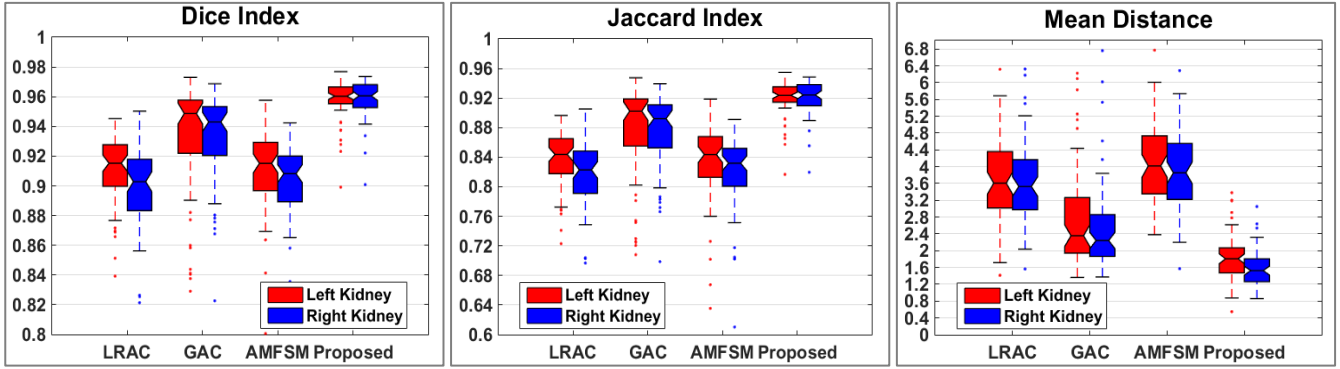

**Fig.14** Segmentation performance measures of segmentation results of US images of 65 subjects obtained by different methods.

**Table 3** Segmentation accuracy measures of results obtained by different methods (Paired Wilcoxon rank sum tests were adopted to compare the proposed method with the alternatives)

|  | Dice Index | | | | Jaccard Index | | | | Mean Distance | | | |
|---|---|---|---|---|---|---|---|---|---|---|---|---|
|  | mean | std | median | p-value | mean | std | median | p-value | mean | std | median | p-value |
| **LRAC** | 0.9051 | 0.0257 | 0.9093 | 4.5e-23 | 0.8276 | 0.0421 | 0.8337 | 4.5e-23 | 3.6527 | 0.9810 | 3.5612 | 4.8e-23 |
| **GAC** | 0.9315 | 0.0374 | 0.9465 | 3.7e-20 | 0.8739 | 00617 | 0.8984 | 3.6e-20 | 2.9888 | 2.0283 | 2.2896 | 2.7e-20 |
| **AMFSM** | 0.9046 | 0.0324 | 0.9111 | 4.5e-23 | 0.8274 | 0.0516 | 0.8367 | 4.5e-23 | 4.0248 | 1.0584 | 3.9098 | 4.5e-23 |
| **Proposed** | 0.9581 | 0.0131 | 0.9604 |  | 0.9204 | 0.0237 | 0.9239 |  | 1.7166 | 0.5043 | 1.6535 |  |

## 4. DISCUSSION AND CONCLUSIONS

In this paper, a dynamic GC based segmentation method with integrated multiple feature maps is proposed to segment kidneys in US images. In particular, the proposed method consists of texture feature extraction and multi-channel GC based segmentation. The feature maps used in our method include the original kidney US image intensity information and the Gabor feature maps. The proposed method has been evaluated on US images of 65 subjects with parameters optimized based on US images of 20 subjects different from the validation dataset. We have also compared our methods with state of the art image segmentation methods, including localizing region based active contours (Lankton and Tannenbaum 2008), geodesic active contours (Caselles, Kimmel et al. 1997), and a recent adaptive multi-feature segmentation model (Zhang, Han et al. 2016). Extensive experiment results have demonstrated that the combination of multiple feature maps can yield better segmentation performance than any single feature map. The proposed method will facilitate development of accurate and reproducible ways to objectively measure kidney features such as echogenicity and cortico-medullary differentiation, which are increased and decreased, respectively, in many kidney diseases. Development of reproducible method to objectively measure these features will lead to identification and validation of anatomic biomarkers that may predict clinically important outcomes such as progression of chronic kidney disease.

The image similarity measure plays a critical role in the GC based segmentation methods. To deal with high speckle noise and low contrast between foreground and background, as well as weak boundaries and large appearance variations in kidney US images, the proposed method integrates multi-feature maps into pixel and localized regional similarity measures to build a narrow band graph. The experimental results have demonstrated that the multi-feature maps based method could yield better performance than any single feature based segmentation.

The proposed method is a semi-automatic segmentation method, and manual initialization is needed. Although the high speckle noise and low contrast in kidney US images may result in instable segmentation results if different initializations are used, good performance can be obtained by constraining the initialization around the kidney boundary and building a narrow band graph. The ICC results also indicated that the segmentation results of the proposed method were robust to different initializations.

We compared our method with stat of the art image segmentation methods, including localizing region based active contours (Lankton and Tannenbaum 2008), geodesic active contours (Caselles, Kimmel et al. 1997), and a recent adaptive multi-feature segmentation model (Zhang, Han et al. 2016). The former two methods are the typical image segmentation methods built upon regional and pixelwise information respectively, and have the best performance among the six methods in the "creaseg" software (Dietenbeck, Alessandrini et al. 2010). The multi-feature based segmentation method (AMFSM) (Zhang, Han et al. 2016) is capable of integrating multiple features using a strategy different from ours. In particular, the AMFSM takes a cosine similarity measure of feature vectors to determine the curve evolution direction and a distance similarity measure to determine the magnitude of driving force, while our method utilize the GC based method to determine the final results by integrating multi-feature maps together and building a narrow band graph, which tends to achieve a global optimal solution. Based on the same image features, our method achieved better segmentation performance than the AMFSM, indicating that our strategy for integrating multiple features is more suitable for Kidney US image segmentation.

In summary, we propose a dynamic localized GC based segmentation method with integrated multiple feature maps to improve the segmentation of Kidney US images. Besides the original image intensity information, Gabor texture feature map is extracted from US images and used as a complementary feature map. The multi-feature maps are finally fused via a novel dynamic localized multi-channel GC. The experimental results demonstrated that the proposed method could achieve promising segmentation performance for segmenting kidneys in US images.

## ACKNOWLEDGEMENTS

This work was supported by National Key Basic Research and Development Program of China (2015CB856404), National Natural Science Foundation of China (61473296), Promotive Research Fund for Excellent Young and Middle-Aged Scientists of Shandong Province (BS2014DX012), China Postdoctoral Science Foundation (2015M581203), The International Postdoctoral Exchange Fellowship Program (20160032), National Institutes of Health grants (EB022573, MH107703, DA039215, and DA039002).

## DISCLOSURE OF CONFLICTS OF INTEREST

The authors have no relevant conflicts of interest to disclose.

## REFERENCES

Ardon, R., R. Cuingnet, K. Bacchuwar and V. Auvray (2015). "Fast kidney detection and segmentation with learned kernel convolution and model deformation in 3D ultrasound image." in: Processing of 12th IEEE International Symposium on Biomedical Imaging (ISBI): 268-271.

Bae, E. and X. C. Tai (2008). "Graph Cuts for the Multiphase Mumford-Shah Model Using Piecewise Constant Level Set Methods." UCLA Applied Mathematics **CAM-report-08-36**

Bernard, O., D. Friboulet, P. Thevenaz and M. Unser (2009). "Variational B-spline level set: A linear filter approach for fast deformable model evolution." IEEE Transactions on Image Processing **18**(6): 1179-1191.

Boykov, Y. and M. Jolly (2001). "Interactive graph cuts for optimal boundary & region segmentation of objects in N-D images." IEEE International Conference on Computer Vision **1**: 105-112.

Boykov, Y. and V. Kolmogorov (2004). "An experimental comparison of min-cut/max-flow algorithms for energy minimization in vision." IEEE Transactions on Pattern Analysis and Machine Intelligence **26**(9): 1124-1137.

Boykov, Y., O. Veksler and R. Zabih (2001). "Fast approximate energy minimization via graph cuts." IEEE Transactions on Pattern Analysis and Machine Intelligence **23**(11): 1222-1239.

Caselles, V., R. Kimmel and G. Sapiro (1997). "Geodesic active contours." International Journal of Computer Vision **22**(1): 61-79.

Cerrolaza, J. J., N. Safdar, E. Biggs and J. Jago (2016). "Renal segmentation from 3D ultrasound via fuzzyappearance models and patient-specific alpha shapes." IEEE Transactions on Medical Imaging 35(11): 1-10.

Chan, T. and L. Vese (2010). "Active contours without edges." IEEE Transactions on Image Processing 10(2): 266-277.

Cohen, F. S. and D. B. Cooper (1987). "Simple parallel hierarchical and relaxation algorithms for segmenting noncausal markovian random fields." IEEE Transactions on Pattern Analysis and Machine Intelligence 9(2): 195-219.

Cost, G. A., P. A. Merguerian, S. P. Cheerasarn and L. M. D. Shortliffe (1996). "Sonographic renal parenchymal and pelvicaliceal areas: New quantitative parameters for renal sonographic followup." Journal of Urology 156(2): 725-729.

Cunha, A. L., J. P. Zhou and M. N. Do (2006). "The nonsubsampled contourlet transform: theory, design, and applications." IEEE Transactions on Image Processing 15(10): 3089-3101.

Daugman, J. G. (1985). "Uncertainty relation for resolution in space, spatial frequency, and orientation optimized by two-dimensional visual cortical filters." Journal of the Optical Society of America. A, Optics and image science 2(7): 1160-1169.

Dietenbeck, T., M. Alessandrini, D. Friboulet and O. Bernard (2010). "Creaseg: a free software for the evaluation of image segmentation algorithms based on level set." IEEE International Conference on Image Processing 119(5): 665-668.

Faucheux, C., J. Olivier, R. Bone and P. Markris (2012). "Texture-based graph regularization process for 2D and 3D ultrasound image segmentation." in 19th IEEE International Conference on Image Processing: 2333-2336.

Gao, J., A. Perlman, S. Kalache, N. Berman, S. Seshan, S. Salvatore, L. Smith, N. Wehrli, L. Waldron, H. Kodali and J. Chevalier (2017). "Multiparametric Quantitative Ultrasound Imaging in Assessment of Chronic Kidney Disease." J Ultrasound Med.

Hacihaliloglu, L., R. Abugharbieh, A. J. Hodgson and R. N. Rohling (2011). "Automatic adaptive parameterization in local phase feature based bone segmentation in Ultrasound." Ultrasound in Medicine and Biology 37(10): 1689-1703.

Haralick, R. M., K. Shanmugam and I. Dinstein (1973). "Textural features for image classification." IEEE Transactions on Systems, Man, and Cybenetics 3(4): 610-621.

Jia, R., S. J. Mellon, S. Hansjee, A. P. Monk, D. W. Murray and J. A. Noble (2016). "Automatic bone segmentation in ultrasound images using loca phase features and dynamic programming." in 2016 IEEE 13th International Symposium on Biomedical Imaging (ISBI): 1005-1008.

Jokar, E. and H. Pourghassem (2013). "Kidney segmentation in Ultrasound images using curvelet transform and shape prior." in 2013 International Conference on Communications Systems and Network Technologies: 180-185.

Jones, J. P. and L. A. Palmer (1987). "An evaluation of the two-dimensional gabor filter model of simple receptive fields in cat striate cortex." Journal of Neurophysiology 58(6): 1233-1258.

Lankton, S. and A. Tannenbaum (2008). "Localizing region based active contours." IEEE Transactions on Image Processing 17(11): 2029-2039.

Le, T. H., S.-W. Jung, K.-S. Choi and S.-J. Ko (2010). "Image segmentation based on modified graph cut algorithm." Electronics Letters 46(16): 1121-1122.

Li, C. M., C. Y. Kao, J. C. Core and D. Z. H. (2008). "Minimization of region-scalable fitting energy for image segmentation." IEEE Transactions on Image Processing 17(10): 1940-1949.

Liu, L. M., W. B. Tao, J. Liu and J. W. Tian (2011). "A variational model and graph cuts optimization for interactive foreground extraction." Signal Processing 91: 1210-1215.

Malik, J., S. Belongie, T. Leung and J. Shi (2001). "Contour and Texture Analysis for Image Segmentation." International Journal of Computer Vision 43(1): 7-27.

Marcelja, S. (1980). "Mathematical description of the responses of simple cortical cells." Journal of the Optical Society of America 70(11): 1297-1300.

Marsousi, M., K. N. Plataniotis and S. Stergiopoulos (2015). "Atlas-based organ segmentation in 3D ultrasound images and its application in automated kidney segmentation." In: Processing of the 37th International Conderence of the IEEE in Engineering in Medicine and Biology Society (EMBC): 2001-2005.

Martin-Fernandez, M. and C. Alberola-Lopez (2005). "An approach for contour detection of human kidneys from ultrasound images using markov random fields and active contours." Medical Image Analysis 9(1): 1-23.

McGraw, K. O. and S. P. Wong (1996). "Forming inferences about some intraclass correlation coefficients." Psychological Methods 1(1): 30-46.

Mehrnaz, Z. Q. and S. Jagath (2009). "2D ultrasound images segmentation using graph cuts and local image features." in IEEE Symposium on Computational Intelligence for Image Processing: 33-40.

Noble, J. A. and D. Boukerroui (2006). "Ultrasound image segmentation: a survey." Medical Imaging, IEEE Transactions on 25(8): 987-1010.

Pulido, J. E., S. L. Furth, S. A. Zderic, D. A. Canning and G. E. Tasian (2014). "Renal Parenchymal Area and Risk of ESRD in Boys with Posterior Urethral Valves." Clinical Journal of the American Society of Nephrology 9(3): 499-505.

Sandberg, B., T. Chan and L. Vese (2002). "A level set and gabor based active contour algorithm for segmenting textured images." UCLA Department of Mathematics CAM report.

Shi, Y. G. and W. Karl (2008). "A real-time algorithm for the approximation of level set based curve evolution." IEEE Transactions on Image Processing **17**(5): 645-656.

Song, Y. H., H. X. Wang, Y. Liu, C. M. Li, G. E. Tasian, Z. X. Gong and D. Z. Zhao (2016). "An improved level set method for segmentation of renal parenchymal area from ultrasound images." Journal of Medical Imaging and Health Informatics **5**(7): 1533-1536.

Tamilselvi, P. R. and P. Thangaraj (2012). "A modified watershed segmentation method to segment renal calculi in ultrasound kidney images." Journal of Intelligent Manufacturing **8**(6): 46-61.

Tao, W. B. (2012). "Iterative narrowband based graph cuts optimization for geodesic active contours with region forces (GACWRF)." IEEE Transactions on Image Processing **21**(1): 284-296.

Tian, Z. Q., L. Z. Liu, Z. F. Zhang and B. W. Fei (2016). "Superpixel based segmentation for 3D prostate MR images." IEEE Transactoins on Medical Imaging **35**(3): 791-801.

Wu, Q. G., Y. Gan, B. Lin, Q. W. Zhang and H. W. Chang (2015). "An active contour model based fused texture features for image segmentation." Neurocomputing **151**(3): 1133-1141.

Xie, J., Y. F. Jiang and H. Tsui (2005). "Segmentation of Kidney form ultrasound images based on texture and shape priors." IEEE Transactions on Medical Imaging **24**(1): 45-57.

Xu, N., N. Ahuja and R. Bansal (2003). "Object segmentation using graph cuts based active contours." in IEEE Conference on Computer Vision & Pattern Recognition (CVPR) **2**(107): 46-53.

Yang, F., W. J. Qin, Y. Q. Xie, T. X. Wen and J. Gu (2012). "A shape optimized framework for kidney segmentation in ultrasound images using NLTV denoising and DRLSE." Biomedical Engineering **11**(1): 1-13.

Zhang, P., Y. M. Liang, S. J. Chang and H. L. Fan (2013). "kidney segmentation in CT sequences using graph cuts based active contours model and contextual continuity." Medical Physics **40**(8): 081905.

Zhang, T. T., J. Han, Y. Zhang and L. F. Bai (2016). "An adaptive multi-feature segmentation model for infrared image." Optical Review **23**: 220-230.